\documentclass[runningheads]{llncs}
\usepackage[utf8]{inputenc}
\usepackage[T1]{fontenc}
\usepackage[pdftex]{graphicx}
\usepackage{amsmath}
\usepackage{amssymb}
\usepackage{bm}
\usepackage{booktabs}
\usepackage{float}
\usepackage{hyperref}
\usepackage{color}

\newcommand{\ourmod}{PANIC}

\begin{document}

\title{Don't PANIC: Prototypical Additive Neural Network for Interpretable Classification of Alzheimer's Disease\thanks{T.N.~Wolf and S.~P{\"{o}}lsterl -- These authors contributed equally to this work.}}

\titlerunning{Prototypical Additive Neural Network for Interpretable Classification}

\author{Tom Nuno Wolf\inst{1,2} \and
Sebastian P{\"o}lsterl\inst{2} \and
Christian Wachinger\inst{1,2}}

\authorrunning{T. N. Wolf et al.}

\institute{
Department of Radiology, Technical University Munich, Munich, Germany \and
Lab for Artificial Intelligence in Medical Imaging, Ludwig Maximilians University, Munich, Germany \email{tom\_nuno.wolf@tum.de}
}

\maketitle

\begin{abstract}
	Alzheimer's disease (AD) has a complex and multifactorial etiology,
	which requires integrating information about neuroanatomy, genetics,
	and cerebrospinal fluid biomarkers for accurate diagnosis.
	Hence, recent deep learning approaches combined image and tabular information
	to improve diagnostic performance.
	However, the black-box nature of such neural networks is still a barrier for clinical applications, in which understanding the decision of a heterogeneous model is integral.
	We propose \ourmod, a prototypical additive neural network for interpretable AD classification that integrates 3D image and tabular data.
	It is interpretable by design and, thus, avoids the need for post-hoc explanations that try to
	approximate the decision of a network.
	Our results demonstrate that \ourmod\ achieves state-of-the-art performance
	in AD classification, while directly providing local and global
	explanations.
	Finally, we show that \ourmod\
	extracts biologically meaningful signatures of AD, and
	satisfies a set of desirable
	desiderata for trustworthy machine learning.
	Our implementation is available at \url{https://github.com/ai-med/PANIC}.
\end{abstract}

\section{Introduction}

It is estimated that the number of people suffering from dementia worldwide will reach
152.8 million by 2050, with Alzheimer's disease (AD) %
accounting for approximately 60--80\% of all cases~\cite{Nichols_2022}.
Due to large studies, like the Alzheimer's Disease Neuroimaging Initiative (ADNI; \cite{ADNI}), and advances in deep learning,
the disease stage of AD can now be predicted relatively accurate~\cite{Wen2020}.
In particular, models utilizing both
tabular and image data have shown performance superior
to unimodal models~\cite{ElSappagh2020,Esmaeil_2018,Wolf_2022_NeuroImage}.
However, they are considered black-box models, as their decision-making process remains largely opaque.
Explaining decisions of Convolutional Neural Networks (CNN)
is typically achieved with post-hoc techniques in the form of saliency maps.
However,
recent studies showed that different post-hoc techniques
lead to vastly different explanations of the same model~\cite{Kindermans_2019}.
Hence, post-hoc methods do not mimic the true model accurately and
have low fidelity~\cite{Rudin_2019}.
Another drawback of post-hoc techniques is that they provide local interpretability
only, i.e., an approximation of the decision of a model for a specific input sample, which cannot explain the
overall decision-making of a model.
Rudin~\cite{Rudin_2019} advocated to overcome these shortcomings
with inherently interpretable models,
which are interpretable by design.
For instance, a logistic regression model is inherently interpretable,
because one can infer the decision-making process from the weights of each feature.
Moreover, inherently interpretable models do provide both local and global
explanations.
While there has been progress towards inherently interpretable
unimodal deep neural networks~(DNNs)~\cite{Rishabh_2021_NEURIPS,Kim_2021_CVPR},
there is a lack of inherently interpretable heterogeneous DNNs
that incorporate both 3D image and tabular data.

\begin{figure}[t]
	\includegraphics[width=\textwidth]{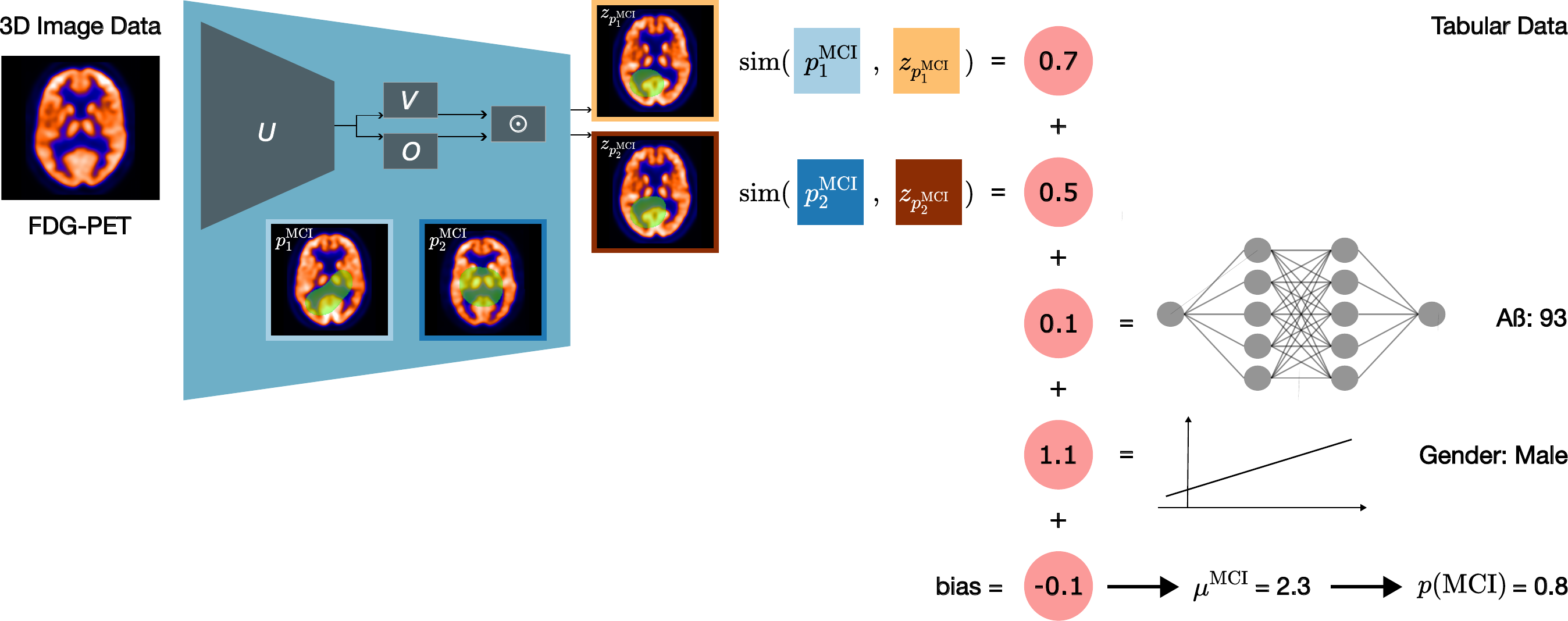}
	\caption{An exemplary prediction for class MCI with \ourmod: 3D FDG-PET images are processed with an interpretable CNN that computes the cosine similarity between latent representations $z_{p^\text{MCI}_1}$,$z_{p^\text{MCI}_2}$ and corresponding prototypes $p^\text{MCI}_1$,$p^\text{MCI}_2$, as seen on the left.
	For each categorical feature, such as gender, a linear function is learned.
	Each continuous feature, such as $\textrm{A}\beta$ is processed with its own MLP.
	The final prediction is the sum of the outputs of the submodules plus a bias term.}
	\label{fig:one}
\end{figure}

In this work, we propose \ourmod, a Prototypical Additive Neural Network for Interpretable Classification of AD,
that is based on the Generalized Additive Model~(GAM).
\ourmod\ consists of one neural net for 3D image data, one neural net for each
tabular feature, and combines their outputs via summation to yield
the final prediction (see Fig.~\ref{fig:one}).
\ourmod\ processes 3D images with an inherently interpretable CNN, as proposed in~\cite{Kim_2021_CVPR}.
The CNN is a similarity-based classifier that
reasons by comparing latent features of an input image
to a set of class-representative latent features.
The latter are representations of specific images from the training data.
Thus, its decision-making can be considered similar to the way humans reason.
Finally, we show that \ourmod\ is fully transparent, because it is interpretable both locally and globally, and achieves
state-of-the-art performance for AD classification.

\section{Related Work}

\paragraph*{Interpretable Models for Tabular Data.}

Decision trees and linear models, such as logistic regression, are inherently interpretable and have been applied widely~\cite{DiMartino_2022}.
In contrast, multi-layer perceptrons (MLPs) are
non-parametric and non-linear, but
rely on post-hoc techniques for explanations.
A GAM is a non-linear model that is fully interpretable, as its
prediction is the sum of the outputs of univariate functions (one for each feature)~\cite{Yin2012}.
Explainable Boosting Machines~(EBMs) extend GAMs by allowing pairwise interaction of features.
While this may boost performance compared to a standard GAM,
the model is harder to interpret, because the number of functions to consider
grows quadratically with the number of features.
EBMs were used in~\cite{Sarica_2021} to predict conversion to AD.

\paragraph*{Interpretable Models for Medical Images.}

ProtoPNet~\cite{Chen_2019_NEURIPS} is a case-based interpretable CNN
that learns class-specific prototypes and defines the prediction as the
weighted sum of the similarities of features, extracted from a given input image,
to the learned prototypes.
It has been applied in the medical domain for diabetic retinopathy grading~\cite{Hesse_2022}.
One drawback is that prototypes are restricted by the size of local patches:
For example, it cannot learn a \textit{single} prototype to represent hippocampal atrophy,
because the hippocampus appears in the left and right hemisphere.
As a result, the number of prototypes needs to be increased to learn a separate prototype for each hemisphere.
The Deformable ProtoPNet~\cite{Donnelly_2022_CVPR} allows for multiple fine-grained prototypical parts
to extract prototypes, but is bound to a fixed number of prototypical parts that represent a prototype.
XProtoNet~\cite{Kim_2021_CVPR} overcomes this limitation by defining prototypes
based on attention masks rather than patches;
it has been applied for lung disease classification from radiographic images.
Wang~et~al.~\cite{Wang_2022_MICCAI} used knowledge distillation to guide the training
of a ProtoPNet for mammogram classification.
However, their final prediction is uninterpretable, because it is the average of the prediction of a ProtoPNet and a black-box CNN.
The works in
\cite{Ilanchezian_2021,Nguyen_2022_MICCAI,Yuki_2021,Wang_2022_MICCAI,Yin_2022_MICCAI}
proposed interpretable models for medical applications, but in contrast to ProtoPNet, XProtoNet and GAMs,
they do not guarantee that explanations are faithful to the model prediction~\cite{Rudin_2019}.

\section{Methods}\label{sec:methods}

We propose a prototypical additive neural network for interpretable classification (\ourmod)
that provides unambiguous local and global explanations for tabular and 3D image data.
\ourmod\ leverages the transparency of GAMs by adding functions that measure similarities between an input image and a set of class-specific prototypes,
that are latent representations of images from the training data~\cite{Kim_2021_CVPR}.

Let the input consist of $N$ tabular features $x_n \in \mathbb{R}$ ($n \in \{1,\dots,N\}$),
and a 3D grayscale image $\mathcal{I} \in \mathbb{R}^{1 \times H \times D \times W}$.
\ourmod\ is a GAM comprising $N$ univariate functions $f_n$ to account
for tabular features, and an inherently interpretable CNN $g$ to
account for image data~\cite{Kim_2021_CVPR}.
The latter provides interpretability by learning a set of $K \times C$ class-specific prototypes
($C$ classes, $K$ prototypes per class $c \in \{1,\ldots,C\}$).
During inference, the model seeks evidence for the presence of prototypical parts in an image,
which can be visualized and interpreted in the image domain.
Computing the similarities of prototypes to latent features representing the presence of prototypical parts
allows to predict the probability of a sample belonging to class $c$:
\begin{equation}\label{eq:ourmod}\textstyle
	p(c\,|\,x_1, \ldots, x_N, \mathcal{I}) = \mathrm{softmax} \left(
		\mu^c
	\right), %
	\quad%
	\mu^c = \beta^c_0 + \sum_{n=1}^N f^c_n(x_n) + \sum_{k=1}^K g^c_k(\mathcal{I}),
\end{equation}
where $\beta_0^c \in \mathbb{R}$ denotes a bias term, $f^c_n(x_n)$ the
class-specific output of a neural additive model for feature $n$,
and $g_k^c(\mathcal{I})$ the similarity between
the $k$-th prototype of class $c$ and the corresponding feature extracted from an input image $\mathcal{I}$.
We define the functions $f_n^c$ and $g_k^c$ below.

\subsection{Modeling Tabular Data}\label{sec:NAM}

Tabular data often consists of continuous
and discrete-valued features, such as age and genetic alterations.
Therefore, we model feature-specific functions $f_n^c$ depending on the type
of feature $n$. If it is continuous, we estimate $f_n^c$
non-parametrically using a multi-layer perceptron~(MLP), as proposed in~\cite{Rishabh_2021_NEURIPS}.
This assures full interpretability while allowing for non-linear processing of each feature $n$.
If feature $n$ is discrete, we estimate $f_n^c$ parametrically using a linear model,
in which case $f_n^c$ is a step function, which is fully interpretable too.
Moreover, we explicitly account for missing values by learning
a class-conditional missing value indicator $s_n^c$.
To summarize, $f^c_n$ is defined as
\begin{equation}\label{eq:fm}
	f^c_n(x_n) = \begin{cases}
		s_n^c, & \text{if $x_n$ is missing,} \\
		\beta_n^c x_n,~\text{with $\beta_n^c \in \mathbb{R}$}, & \text{if $x_n$ is categorical} \\
		\mathrm{MLP}_n^c (x_n), & \text{otherwise.}
	\end{cases}
\end{equation}
Predicting a class with the sum of such univariate functions $f^c_n$ was proposed in~\cite{Rishabh_2021_NEURIPS} as Neural Additive Model~(NAM).
Following~\cite{Rishabh_2021_NEURIPS}, we apply an $\ell_2$ penalty on the outputs of $f^c_n(x_n)$:
\begin{equation*}\label{eq:nam-l2-penalty}\textstyle
	\mathcal{L}_\text{Tab}(x_1,\ldots,x_n) = \frac{1}{C} \sum_{c=1}^C \sum_{n=1}^N [f^c_n(x_n)]^2.
\end{equation*}

\begin{figure}[t]
	\includegraphics[width=\textwidth]{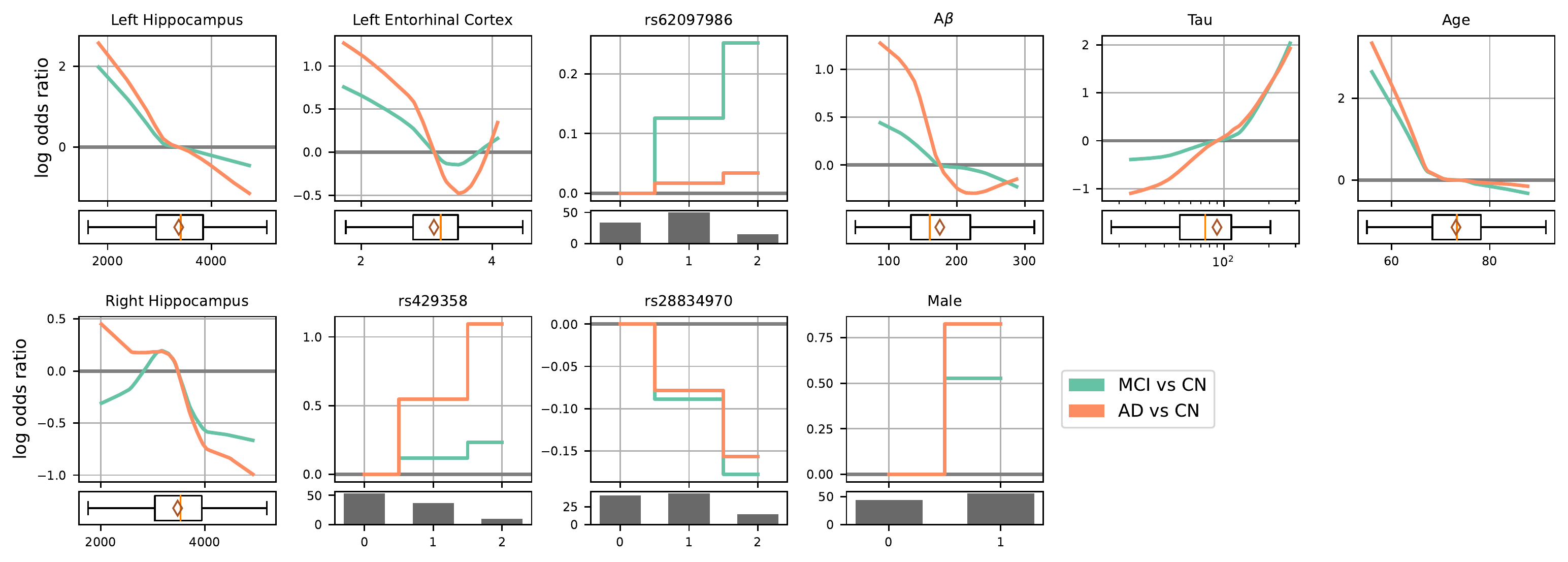}
	\caption{\label{fig:nam_plots}%
	The plots show the log odds ratio with respect to controls
	for the top 10 tabular features.
	Boxplots show the distribution in our data.
	}
\end{figure}

\begin{figure}[t]
	\includegraphics[width=\textwidth]{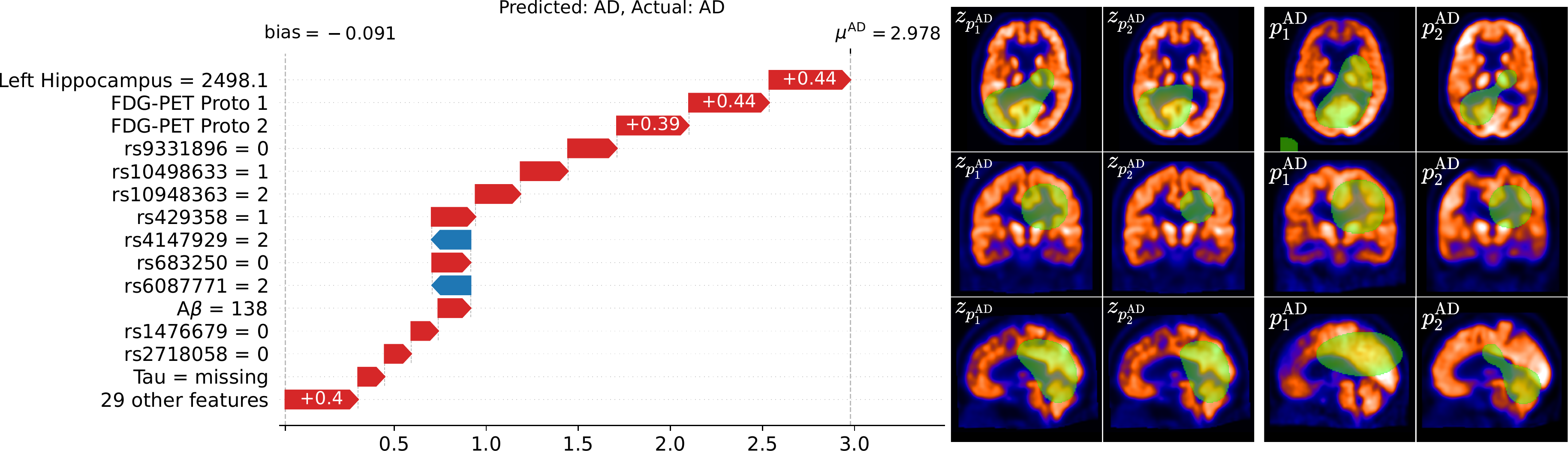}
	\caption{\label{fig:pred_waterfall}%
	Explanation for the prediction of a single sample from the test set.
	Left: Individual contributions to the overall prediction.
	Right: Input FDG-PET overlayed with the attention maps (green) of the corresponding representations $z_{p^\text{AD}_1}$,$z_{p^\text{AD}_1}$ (columns 1,2), and the attention maps of learned prototypes $p^\text{AD}_1$,$p^\text{AD}_2$ (columns 3,4).
	}
\end{figure}

We want to emphasize that NAMs retain global interpretability by plotting each univariate function $f^c_n$
over its domain (see Fig.~\ref{fig:nam_plots}).
Local interpretability
is achieved by evaluating $f^c_n(x_n)$,
which equals the contribution of a feature $x_n$ to the prediction of a sample, as defined
in equation~\eqref{eq:ourmod}
(see Fig.~\ref{fig:pred_waterfall} on the left). %

\subsection{Modeling Image Data}\label{sec:XProtoNet}

We model 3D image data by defining the function
$g_k^c(\mathcal{I})$ in equation~\eqref{eq:ourmod}
based on XProtoNet~\cite{Kim_2021_CVPR}, which
learns prototypes that can span multiple, disconnected regions within an image.
In XProtoNet,
an image is classified based on the cosine similarity between a latent feature vector $z_{p^c_k}$
and learned class-specific prototypes $p^c_k$, as depicted in the top part of Fig.~\ref{fig:one}:
\begin{equation}\label{eq:xprotonet-similarity}\textstyle
	g^c_k(\mathcal{I}) = \mathrm{sim}(p^c_k, z_{p^c_k}) = \frac{ p^c_k \cdot z_{p^c_k} }{ \lVert p^c_k \rVert \lVert z_{p^c_k} \rVert }.
\end{equation}
A latent feature vector $z_{p^c_k}$ is obtained by passing an image $\mathcal{I}$
into a CNN backbone $\mathcal{U}:\mathbb{R}^{1 \times H \times D \times W} \rightarrow \mathbb{R}^{R \times H^\prime \times D^\prime \times W^\prime}$,
where $R$ is the number of output channels.
The result is passed into two separate modules:
(i) the feature extractor
$\mathcal{V}: \mathbb{R}^{R \times H^\prime \times D^\prime \times W^\prime} \rightarrow \mathbb{R}^{L \times H^\prime \times D^\prime \times W^\prime}$
maps the feature map to the dimensionality of the prototype space $L$;
(ii) the occurrence module
$\mathcal{O}^c: \mathbb{R}^{R \times H^\prime \times D^\prime \times W^\prime} \rightarrow \mathbb{R}^{K \times H^\prime \times D^\prime \times W^\prime}$
produces $K$ class-specific attention masks.
Finally, the latent feature vector $z_{p^c_k}$ is defined as
\begin{equation}
	z_{p^c_k} = \textrm{GAP} [
		\textrm{sigmoid}(\mathcal{O}^c(\mathcal{U}(\mathcal{I}))_k)
		\odot \mathrm{softplus}(\mathcal{V}(\mathcal{U}(\mathcal{I})))
	],
\end{equation}
where $\odot$ denotes the Hadamard product, and GAP global average pooling.

Intuitively, $z_{p^c_k}$ represents the GAP-pooled activation maps that a prototype $p^c_k$ would yield if it were present in that image.  %
For visualization, we can upsample the occurrence map $\mathcal{O}^c(\mathcal{U}(\mathcal{I}))_k$ to the input dimensions and overlay it on the input image.
The same can be done to visualize prototype $p^c_k$
(see Fig~\ref{fig:pred_waterfall}).

\paragraph*{Regularization.}
Training XProtoNet requires regularization with respect to the occurrence module and prototype space~\cite{Kim_2021_CVPR}:
An occurrence and affine loss enforce sparsity and spatial fidelity of the attention masks $\mathcal{O}^c$ with respect to the image domain:
\begin{equation*}\textstyle
	\mathcal{L}_\text{occ}(\mathcal{I}) =
		\sum_{c=1}^C
		\lVert \mathcal{O}^c(\mathcal{U}(\mathcal{I})) \rVert_1, \quad
	\mathcal{L}_\text{affine}(\mathcal{I}) = \lVert A(\mathcal{O}^c(\mathcal{U}(\mathcal{I}))) - \mathcal{O}^c(\mathcal{U}(A(\mathcal{I})))
		\rVert_1,
\end{equation*}
with $A$ a random affine transformation.
Additionally, latent vectors $z_{p^c_k}$ of an
image $\mathcal{I}$ with true class label $y$
should be close to prototypes of their
respective class, and distant to prototypes of other classes:
\begin{equation*}\textstyle
	\mathcal{L}_\text{clst}(\mathcal{I}) = -\max_{k, c=y}~ g^c_k(\mathcal{I}), \quad
	\mathcal{L}_\text{sep}(\mathcal{I}) = \max_{k, c \neq y}~ g^c_k(\mathcal{I}) .
\end{equation*}

\subsection{\ourmod}

As stated in equation~\eqref{eq:ourmod}, \ourmod\ is a GAM comprising
functions $f_1^c,\ldots,f_N^c$ for tabular data,
and functions $g_1^c,\ldots,g_K^c$ for 3D image data
(see equations \eqref{eq:fm} and \eqref{eq:xprotonet-similarity}).
Tabular features contribute to the overall prediction in equation~\eqref{eq:ourmod}
in terms of the values
$f_1^c(x_1),\ldots,f_N^c(x_N)$, while
the image contributes in terms of the cosine similarity between the class-specific prototype
$p^c_k$ and the latent feature vector $z_{p^c_k}$.
By restricting prototypes to contribute to the prediction of a specific class only,
we encourage the model to learn discriminative prototypes for each class.

To interpret \ourmod\ locally,
we simply consider the outputs of the functions $f_n^c$,
and sum the image-based similarity scores over all prototypes:
$\sum_{k=1}^K g_k^c(\mathcal{I})$.
To interpret the contributions due to the 3D image in detail,
we examine the attention map of each prototype,
the attention map of the input image,
and the similarity score between each prototype and the image
(see Fig.~\ref{fig:pred_waterfall} on the right).
To interpret \ourmod\ globally, we compute the absolute
contribution of each function to the per-class logits
in equation~\eqref{eq:ourmod}, and average it over
all samples in the training set, as seen in Fig.~\ref{fig:globalimportance}.
In addition, we can directly visualize the function $f_n^c$
learned from the tabular data in terms of the log odds ratio
$$
\log \left[ \frac{p(c\,|\, x_1, \ldots, x_n, \ldots, x_N, \mathcal{I})}
            {p(\text{CN}\,|\, x_1, \ldots, x_n, \ldots, x_N, \mathcal{I})}
\big/ \frac{p(c\,|, x_1, \ldots, x_n^\prime, \ldots, x_N, \mathcal{I})}
           {p(\text{CN}\,|\, x_1, \ldots, x_n^\prime, \ldots, x_N, \mathcal{I})}
\right] ,
$$
where $x_n^\prime$ is the mean value of feature $n$ across
all samples for continuous features, and zero for categorical features.
As an example, let us consider the AD class.
If the log odds ratio for a specific value $x_n$ is positive,
it indicates that the odds of being diagnosed as AD, compared to CN, increases.
Conversely, if it is negative, the odds of being diagnosed as AD decreases.

We train \ourmod\ with the following loss:
\begin{equation}\begin{split}\label{eq:loss}
	\mathcal{L}(y, x_1,\ldots,x_n, \mathcal{I}) &=
	\mathcal{L}_\text{CE}(y, \hat{y})
	+ \lambda_1 \mathcal{L}_\text{Tab}(x_1,\ldots,x_n)
	+ \lambda_2 \mathcal{L}_\text{clst}(\mathcal{I}) \\
	&\quad+ \lambda_3 \mathcal{L}_\text{sep}(\mathcal{I})
	+ \lambda_4 \mathcal{L}_\text{occ}(\mathcal{I}) + \lambda_5 \mathcal{L}_\text{affine}(\mathcal{I}), \\
	\hat{y} &= \arg\max_c ~ p(c\,|\, x_1,\ldots,x_n, \mathcal{I}),
\end{split}\end{equation}
where $\mathcal{L}_\text{CE}$ is the cross-entropy loss,
$\lambda_{1,\dots,5}$ are hyper-parameters,
$y$ the true class label, and $\hat{y}$ the prediction of \ourmod.

\section{Experiments}

\subsection{Overview}
\subsubsection{Dataset.}
Data used in this work was obtained from the ADNI database.\footnote{\url{https://adni.loni.usc.edu/}~\cite{ADNI}}
We select only baseline visits to avoid data leakage, and
use FDG-PET images following the processing pipeline described in~\cite{Narazani2022}.
Tabular data consists of the continuous features age, education;
the cerobrospinal fluid markers $\textrm{A}\beta$, Tau, p-Tau;
the MRI-derived volumes of the left/right hippocampus and
thickness of the left/right entorhinal cortex.
The categorical features are gender and 31 AD-related
genetic variants identified in~\cite{Hibar2015,Lambert2013}
and having a minor allele frequency of $\geq 5\%$.
Tabular data was standardized using the mean and standard deviation
across the training set.
Table~\ref{tab:datadistribution} summarizes our data.

\begin{table}[t]
	\centering
	\caption{\label{tab:datadistribution}%
	Statistics for the data used in our experiments.}
	\begin{scriptsize}
	\begin{tabular*}{\textwidth}{@{\extracolsep{\fill}}lllll}
	  \toprule
	 & CN (N=379) & Dementia (N=256)  & MCI (N=610) & Total (N=1245) \\ %
	  \midrule
	\textbf{Age} &  &  &  &  \\ %
	Mean (SD) & 73.5 (5.9) & 74.5 (7.9) & 72.3 (7.3) & 73.1 (7.1) \\ %
		Range & 55.8 - 90.1 & 55.1 - 90.3 & 55.0 - 91.4 & 55.0 - 91.4 \\[.2em] %
	\textbf{Gender} &  &  &  &  \\ %
	Female & 193 (50.9\%) & 104 (40.6\%) & 253 (41.5\%) & 550 (44.2\%) \\ %
		Male & 186 (49.1\%) & 152 (59.4\%) & 357 (58.5\%) & 695 (55.8\%) \\[.2em] %
	\textbf{Education} &  &  &  &  \\ %
	Mean (SD) & 16.4 (2.7) & 15.4 (2.8) & 16.1 (2.7) & 16.0 (2.8) \\ %
		Range & 7.0 - 20.0 & 4.0 - 20.0 & 7.0 - 20.0 & 4.0 - 20.0 \\[.2em] %
	\textbf{MMSE} &  &  &  & \\ %
	Mean (SD) & 29.0 (1.2) & 23.2 (2.2) & 27.8 (1.7) & 27.2 (2.7) \\ %
	Range & 24.0 - 30.0 & 18.0 - 29.0 & 23.0 - 30.0 & 18.0 - 30.0 \\ %
	   \bottomrule
	\end{tabular*}
	\end{scriptsize}
\end{table}

\subsubsection{Implementation Details.}

We train \ourmod\ with the loss in equation~\eqref{eq:loss} with AdamW~\cite{AdamW} and a cyclic learning rate scheduler~\cite{CycleLR},
with a learning rate of 0.002, and weight decay of 0.0005.
We choose a 3D ResNet18 for the CNN backbone $\mathcal{U}$ with $R = 256$ channels in the last ResBlock.
The feature extractor $\mathcal{V}$ and the occurrence module $\mathcal{O}$ are CNNs consisting of $1 \times 1 \times 1$ convolutional layers with ReLU activations.
We set $K = 2$, $L = 64$, $\lambda_1 = 0.01$ and $\lambda_{2,\dots,5} = 0.5$,
and norm the length of each prototype vector to one.
For each continuous tabular feature $n$, $f_n^c$ is a MLP
that shares parameters for the class dependent outputs
in \eqref{eq:fm}.
Thus, each MLP has 2 layers with 32 neurons, followed by
an output layer with $C$ neurons.
As opposed to \cite{Rishabh_2021_NEURIPS}, we found it helpful to replace the ExU activations by ReLU activations.
We apply spectral normalization~\cite{Miyato_2018} to make MLPs Lipschitz continuous.
We add dropout with a probability of 0.4 to MLPs, and with a probability of 0.1 to all univariate functions in equation~\eqref{eq:ourmod}.
The set of affine transformations $A$ comprises all transformations
with a scale factor $\in [0.8; 1.2]$
and random rotation $\in [-180^\circ; 180^\circ]$ around the origin.
We initialize the weights of $\mathcal{U}$ from a pre-trained model
that has been trained on the same training data on the classification task for 100 epochs with early stopping.
We cycle between optimizing all parameters of the network and optimizing
the parameters of $f_n^c$ only.
We only validate the model directly after prototypes $p^c_k$ have been replaced
with their closest latent feature vector $z_{p^c_k}$ of samples
from the training set of the same class.
Otherwise, interpretability on an image level would be lost.

We perform 5-fold cross-validation, based on a data stratification strategy that accounts for sex, age and diagnosis.
Each training set is again split such that 64\% remain training and 16\% are used for hyper-parameter tuning (validation set).
We report the mean and standard deviation of the balanced accuracy (BAcc) of the best  hyper-parameters found on the validation sets.

\subsection{Classification Performance}
\ourmod\ achieves 64.0$\pm$4.5\% validation BAcc and 60.7$\pm$4.4\% test BAcc.
We compare \ourmod\ to a black-box model for heterogeneous data, namely DAFT~\cite{Wolf_2022_NeuroImage}.
We carry out a random hyper-parameter search with 100 configurations for learning rate, weight decay and the bottleneck factor of DAFT.
DAFT achieves a validation and test BAcc of 60.9$\pm$0.7\% and 56.2$\pm$4.5\%, respectively.
This indicates that interpretability does not necessitate a loss in prediction performance.

\subsection{Interpretability}
\begin{figure}[t]
	\includegraphics[width=0.75\textwidth]{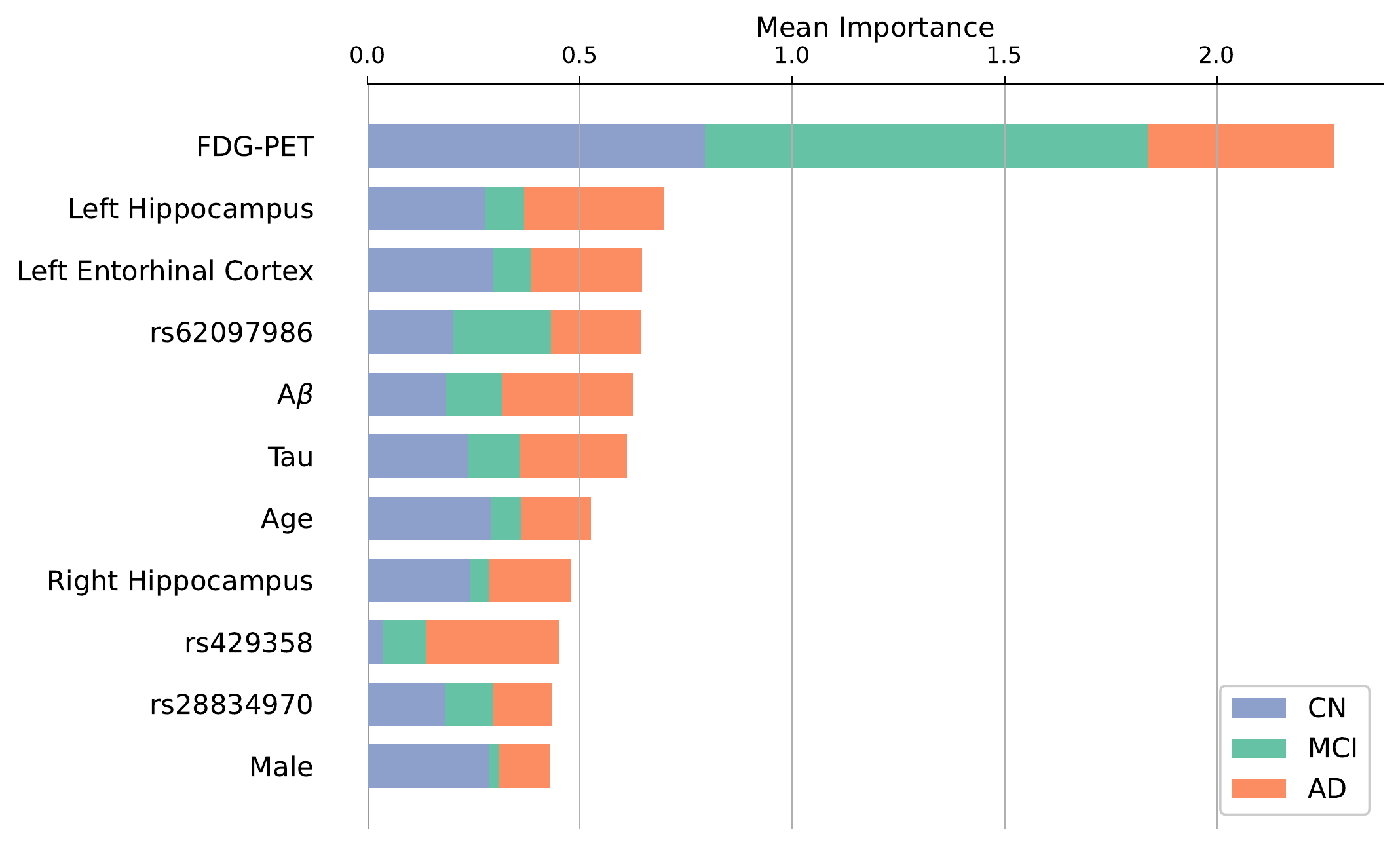}
	\caption{\label{fig:globalimportance}%
	Ranking of the most import features learned by \ourmod.
	FDG-PET denotes the combined importance of all prototypes related to the FDG-PET image.}
\end{figure}

\ourmod\ is easy to interpret on a global and local level.
Figure~\ref{fig:globalimportance} summarizes the average
feature importance over the training set.
It shows that FDG-PET has on average the biggest influence
on the prediction, but also that importance can vary across
classes. For instance, the SNP rs429358, which is located in the ApoE gene,
plays a minor role for the controls, but is highly important
for the AD class. This is reassuring, as it is a well known
risk factor in AD~\cite{Scheltens_2016}.
The overall most important SNP is rs62097986, which is suspected to
influence brain volumes early in neurodevelopment~\cite{Hibar2015}.

To get a more detailed inside into \ourmod,
we visualize the log odds ratio with respect to
the function $f_n^c$ across the domain
of the nine most important tabular features
in Fig.~\ref{fig:nam_plots}.
We can easily see that \ourmod\ learned
that atrophy of the left hippocampus
increases the odds of being diagnosed as AD.
The volume of the right hippocampus is utilized similarly.
For MCI, it appears as \ourmod\ has overfit on outliers
with very low right hippocampus volume.
Overall, the results for left/right hippocampus
agree with the observation that the hippocampus is among the
first structures experiencing neurodegeneration in AD~\cite{Sabuncu_2011}.
The function for left entorhinal cortex thickness agrees with
previous results too~\cite{Sabuncu_2011}.
An increase in $\text{A}\beta$ measured in CSF is associated with
a decreased risk of AD~\cite{Scheltens_2016}, which our model captured correctly.
The inverse relationship holds for Tau~\cite{Scheltens_2016}, which \ourmod\ confirmed too.
The function learned for age shows a slight decrease in the log odds ratio
of AD and MCI, except for patients around 60 years of age, which
is due to few data samples for this age range in the
training data.
We note that the underlying causes that explain the evolution
from normal aging to AD remain unknown, but since age is
considered a confounder one should control for it~\cite{Jagust_2018}.
Overall, we observe that \ourmod\ learned a highly non-linear
function for the continuous features hippocampus volume,
entorhinal thickness, $\text{A}\beta$, Tau, and age,
which illustrates that estimating the functions $f_n^c$ via MLPs is effective.
In our data, males have a higher incidence of AD (see Tab.~\ref{tab:datadistribution}),
which is reflected in the decision-making of the model too.
Our result that rs28834970 decreases the odds for AD
does not agree with previous results~\cite{Lambert2013}.
However, since \ourmod\ is fully interpretable, we can easily
spot this misconception.

Additionally, we can visualize the prototypes by upscaling
the attention map specific to each prototype, as produced by the occurrence module,
to the input image size and highlighting activations of more than
30\% of the maximum value, as proposed in~\cite{Kim_2021_CVPR}
(see Fig.~\ref{fig:pred_waterfall} on the right).
The axial view of $p_1^\text{AD}$ shows attention towards the occipital lobe and $p_2^\text{AD}$ towards one side of the occipital lobe.
Atrophy around the ventricles can be seen in FDG-PET~\cite{Scheltens_2016} and both prototypes $p_1^\text{AD}$ and $p_2^\text{AD}$ incorporate this information, as seen in the coronal views.
The sagittal views show, that $p_2^\text{AD}$ focuses on the cerebellum and parts of the occipital lobe.
The parietal lobe is clearly activated by the prototype in the sagittal view of $p_1^\text{AD}$ and was linked to AD previously~\cite{Scheltens_2016}.

We can interpret the decision-making of \ourmod\ for a specific subject
by evaluating the contribution of each function with respect to the
prediction (see Fig.~\ref{fig:pred_waterfall} on the left).
The patient was correctly classified as AD, and most evidence
supports this (red arrows).
The only exceptions are the SNPs rs4147929 and rs6087771,
which the models treats as evidence against AD.
Hippocampus volume contributed most to the prediction, i.e.,
was most important.
Since, the subject's left hippocampus volume is relatively low
(see Fig.~\ref{fig:nam_plots}), this increases our
trust in the model's prediction.
The subject is heterozygous for rs429358 (ApoE), a
well known marker of AD, which the model captured correctly~\cite{Scheltens_2016}.
The four variants rs9331896, rs10498633, rs4147929, and rs4147929
have been associated with nucleus accumbens volume~\cite{Hibar2015}, which
is involved in episodic memory function~\cite{Yi2015}.
Atrophy of the nucleus accumbens is
associated with cognitive impairment~\cite{Yi2015}.
FDG-PET specific image features present in the image
show a similarity of 0.44 to the features of prototype $p_1^{\text{AD}}$.
It is followed by minor evidence of features similar to prototype $p_2^{\text{AD}}$.
During the prediction of the test subject, the network extracted prototypical parts from similar regions.
As seen in the axial view, both parts $z_{p_1^\text{AD}}$ and $z_{p_2^\text{AD}}$ contain parts of the occipital lobe.
Additionally, they cover a large part of the temporal lobe, which has been linked to AD~\cite{Scheltens_2016}.

In summary, the decision-making of \ourmod\ is easy to comprehend and predominantly agrees with current knowledge about AD.

\section{Desiderata for Machine Learning Models}

We now show that \ourmod\ satisfies
four desirable desiderata for machine learning (ML)
models, based on the work in~\cite{Rudin_2019}.

\textbf{Explanations must be faithful to the underlying model (perfect fidelity).}
To avoid misconception, an explanation must not mimic the underlying model,
but equal the model.
The explanations provided by \ourmod\ are the values provided by the functions
$f_1^c,\ldots,f_N^c,g_1^c,\ldots,g_K^c$ in equation~\eqref{eq:ourmod}.
Since \ourmod\ is a GAM, the sum of these values (plus bias) equals
the prediction. Hence, explanations of \ourmod\ are faithful to how
the model arrived at a specific prediction.
We can plot these values to gain local interpretability, as done in Fig.~\ref{fig:pred_waterfall}.

\textbf{Explanations must be detailed.}
An explanation must be comprehensive such that it provides all information
about what and how information is used by a model (global interpretability).
The information learned by \ourmod\ can be described precisely.
Since \ourmod\ is based on the sum of univariate functions,
we can inspect individual functions to understand what the model learned.
Plotting the functions $f_n^c$ over their domain
explains the change in odds when the value $x_n$ changes,
as seen in Fig.~\ref{fig:nam_plots}.
For image data, \ourmod\ uses the similarity between the features
extracted from the input image and the $K$ class-specific prototypes.
Global interpretability is achieved by visualizing the training image
a prototype was mapped to, and its corresponding attention map~\ref{fig:pred_waterfall}.

\textbf{A machine learning model should help to improve the knowledge discovery process.}
For AD, the precise cause of cognitive decline remains elusive.
Therefore, ML should help to identify biomarkers and relationships,
and inform researchers studying the biological causes of AD.
\ourmod\ is a GAM, which means it provides full global interpretability.
Therefore, the insights it learned from data are directly accessible,
and provide unambiguous feedback to the knowledge discovery process.
For instance, we can directly infer what \ourmod\ learned about
FDG-PET or a specific genetic mutation,
as in Figs.~\ref{fig:nam_plots} and~\ref{fig:globalimportance}.
This establishes a feedback loop connecting ML researchers
and researchers studying the biological causes of AD,
which will ultimately make diagnosis more accurate.

\textbf{ML models must be easy to troubleshoot.}
If an ML model produces a wrong or unexpected result,
we must be able to easily troubleshoot the model.
Since \ourmod\ provides local and global interpretability,
we can easily do this.
We can use a local explanation
(see Fig.~\ref{fig:pred_waterfall}) to precisely
determine what the deciding factor for the prediction was
and whether it agrees with our current understanding of AD.
If we identified the culprit,
we can inspect
the function $f_n^c$, in the case of a tabular feature,
or the prototypes, in the case of the image data:
Suppose, the age of the patient in Fig.~\ref{fig:pred_waterfall} was falsely
recorded as 30 instead of 71.
The contribution of age $f_\text{age}^\text{AD}$ would increase from $-0.138$ to
$3.28$, thus, dominate the prediction by a large margin.
Hence, the local explanation would reveal that something is amiss
and prompt us to investigate the learned function $f_\text{age}^\text{AD}$ (see Fig.~\ref{fig:nam_plots}),
which is ill-defined for this age.

\section{Conclusion}

We proposed an inherently interpretable neural network for tabular and 3D image data, and showcased its use for AD classification.
We used local and global interpretability properties of \ourmod\ to verify that the decision-making of our model largely agrees with current knowledge about AD, and is easy to troubleshoot.
Our model outperformed a state-of-the-art black-box model and satisfies a set of desirable desiderata that establish trustworthiness in \ourmod.

\subsubsection{Acknowledgements}

This research was partially supported by the Bavarian State Ministry of Science and the Arts and coordinated by the bidt, the BMBF  (DeepMentia, 031L0200A), the DFG and the LRZ.

\bibliographystyle{splncs04}
\bibliography{panic}

\end{document}